\documentclass{article}
\usepackage[preprint]{neurips_2025}

\usepackage[utf8]{inputenc} % allow utf-8 input
\usepackage[T1]{fontenc}    % use 8-bit T1 fonts
\usepackage{hyperref}       % hyperlinks
\usepackage{url}            % simple URL typesetting
\usepackage{booktabs}       % professional-quality tables
\usepackage{amsfonts}       % blackboard math symbols
\usepackage{nicefrac}       % compact symbols for 1/2, etc.
\usepackage{microtype}      % microtypography
\usepackage{xcolor}         % colors
\usepackage{graphicx}
\usepackage{multirow}
\usepackage{multicol}
\usepackage{tabularx}
\usepackage{array}
\usepackage{multicol}
\usepackage{makecell}
\usepackage{adjustbox}
\usepackage{siunitx}
\usepackage{arydshln}
\usepackage{adjustbox} % 必需: 支持表格调整和缩放
\usepackage{booktabs} % 必需: 高级表格排版
\usepackage{arydshln} % 必需: 支持虚线
\title{ScaleTrack: Scaling and back-tracking Automated GUI Agents}
\author{
Jing Huang\textsuperscript{\rm 1\thanks{Equal contribution.}},
Zhixiong Zeng\textsuperscript{\rm 1\footnotemark[1]\; \thanks{Corresponding author.}},
Wenkang Han\textsuperscript{\rm 1,2},
Yufeng Zhong\textsuperscript{\rm 1}, \\
\textbf{Liming Zheng}\textsuperscript{\rm 1},
\textbf{Shuai Fu}\textsuperscript{\rm 3},
\textbf{Jingyuan Chen}\textsuperscript{\rm 2},
\textbf{Lin Ma}\textsuperscript{\rm 1\footnotemark[2]}
\\
\textsuperscript{1}\small Meituan \\
\textsuperscript{2}\small Zhejiang University \\
\textsuperscript{3}\small University of Adelaide \\
}

\begin{document}
\maketitle
% \renewcommand{\thefootnote}{}
% \footnotetext{*Equal contribution  $^\dagger$Corresponding authors.}

\begin{abstract}
Automated GUI agents aims to facilitate user interaction by automatically performing complex tasks in digital environments, such as web, mobile, desktop devices.
It receives textual task instruction and GUI description to generate executable actions (\emph{e.g.}, click) and operation boxes step by step. 
Training a GUI agent mainly involves grounding and planning stages, in which the GUI grounding focuses on finding the execution coordinates according to the task, while the planning stage aims to predict the next action based on historical actions.
However, previous work suffers from the limitations of insufficient training data for GUI grounding, as well as the ignorance of back-tracking historical behaviors for GUI planning.
To handle the above challenges, we propose ScaleTrack, a training framework by scaling grounding and back-tracking planning for automated GUI agents.
We carefully collected GUI samples of different synthesis criterions from a wide range of sources, and unified them into the same template for training GUI grounding models.
Moreover, we design a novel training strategy that predicts the next action from the current GUI image, while also back-tracking the historical actions that led to the GUI image. In this way, ScaleTrack explains the correspondence between GUI images and actions, which effectively describes the evolution rules of the GUI environment.
Extensive experimental results demonstrate the effectiveness of ScaleTrack. Data and code will be available at url.
\end{abstract}
\section{Introduction}
\label{sec:intro}

% \begin{figure}[t]
%   \centering
%   \fbox{\rule{0pt}{2in} \rule{0.9\linewidth}{0pt}}
%    %\includegraphics[width=0.8\linewidth]{egfigure.eps}

%    \caption{Example of caption.
%    It is set in Roman so that mathematics (always set in Roman: $B \sin A = A \sin B$) may be included without an ugly clash.}
%    \label{fig:onecol}
% \end{figure}

% \begin{figure*}
%   \centering
%   \begin{subfigure}{0.68\linewidth}
%     \fbox{\rule{0pt}{2in} \rule{.9\linewidth}{0pt}}
%     \caption{An example of a subfigure.}
%     \label{fig:short-a}
%   \end{subfigure}
%   \hfill
%   \begin{subfigure}{0.28\linewidth}
%     \fbox{\rule{0pt}{2in} \rule{.9\linewidth}{0pt}}
%     \caption{Another example of a subfigure.}
%     \label{fig:short-b}
%   \end{subfigure}
%   \caption{Example of a short caption, which should be centered.}
%   \label{fig:short}
% \end{figure*}

GUI agent aims to develop native automated agents for the Graphical User Interface (GUI), and satisfies the growing demands for users to automatically perform complex tasks in digital environments.
It has attracted widespread attention in the area of mobile/compute use, which can provide accurate task completion and convenient user interaction.
Benefiting from the emerging reasoning capabilities of large language models~\cite{team2024gemma,achiam2023gpt}, GUI agents are capable of reasoning and planning multi-step executable actions from task instructions.

Early GUI agents \cite{kim2023language,zheng2023synapse} extract structured text describing the GUI environment (\emph{e.g.}, website HTML), and then leverage large language models to reason and plan for generating
 executable actions. In fact, the structured text information of GUI environment is often lengthy and may not be accessible, and is accompanied by insufficient attribute information such as location and layout. Therefore, recent research \cite{cheng2024seeclick,wu2024atlas} focuses on visual GUI agents based on multimodal large language models (MLLMs), which only rely on the screenshots of the GUI environment for user interaction.

 To train GUI agents based on visual screenshot, existing methods \cite{xu2024aguvis,qin2025ui} typically follow the paradigm of transfering the general multimodal knowledge from MLLMs (\emph{e.g.}, QWen2-VL) to the GUI environment.
 Typically, previous work usually fine-tunes MLLMs on two types of GUI data: GUI grounding for predicting coordinate position and GUI planning for predicting actions.
 The former focuses on predicting the associated coordinate position according to the task instruction, while the latter focuses on predicting the multi-step executable actions, and finally realizing the automated task execution.

 However, existing GUI grounding methods mainly rely on isolated data synthesis criterion to generate grounding data from a large amount of metadata including mobile, web and desktop. 
 Specifically, they adopt a strong LLM (\emph{e.g.}, GPT-4o) that takes the metadata of user interfaces (\emph{e.g.}, metadata of all texts/icons/widgets) on different platforms to synthesize element descriptions.
 Several methods emphasize the isolated data synthesis criterion, in which Uground \cite{gou2024navigating} synthesizes grounding data with multiple reference modes, Aria-UI \cite{yang2024aria} synthesizes context-aware grounding data, and Augvis \cite{xu2024aguvis} synthesizes template-enhanced grounding data. 
 Unfortunately, previous work ignores the complementarity of grounding data with different synthesis methods, thus rarely integrates them for scaling the training process of GUI grounding.

 Futhermore, existing research only considers forward-planning to predict next action based on task instruction and GUI environment. It typically relies on human annotation to collect action trajectory for various task instructions \cite{deng2023mind2web}, or prompts MLLMs to generate detailed reasoning thoughts \cite{xu2024aguvis}.
 In fact, the GUI environment follows intrinsic patterns inherent to task execution, such as forward-planning the next actions that can be taken in the current state, or back-tracing the historical actions that led to the current state.
 However, existing work only collects forward-planning data during training, lacking the collection of back-tracking data as well as the exploration of corresponding training strategies.

 In this paper, we introduce ScaleTrack, a novel training framework by scaling GUI grounding and back-tracking GUI planning to handle the above challenges.
 First, we utilize several data-driven GUI element enhancement methods to scale the training process of GUI grounding, including element referring, context awareness, and functional description.
 To this end, our work integrates a wide range of grounding samples generated from isolated data synthesis criterion and unifies them into a fixed training template.
 Then we designed a data template that integrates the action annotations between multiple consecutive GUI images, and collects the next action groundtruth for forward-planningg and the historical action groundtruth for back-tracking (as shown in Figure 1). 
 Finally, to learn the inherent patterns of task execution, we design a hybrid training strategy includes forward-planning and back-tracking, in which the GUI agent is required to predict the next action of the current state and the historical actions simultaneously.
 Empirical results in the experiment indicate that back-tracking effectively improves the task execution of GUI agents.

The main contributions of this work are as follows:
\begin{itemize}
    \item We propose the first GUI agent with back-tracking capability and devise effective data construction as well as training strategy.
    \item We integrate several data synthesis criterions to enrich GUI element description, significantly scaling the training process that leads to consistent performance improvements.
    \item We conduct extensive experiments on several benchmark datasets including grounding evaluation, offline and online evaluation, and the experimental results verify the effectiveness of our proposed ScaleTrack.
\end{itemize}

\begin{figure}[t]
\centering
\includegraphics[width=0.65\textwidth]{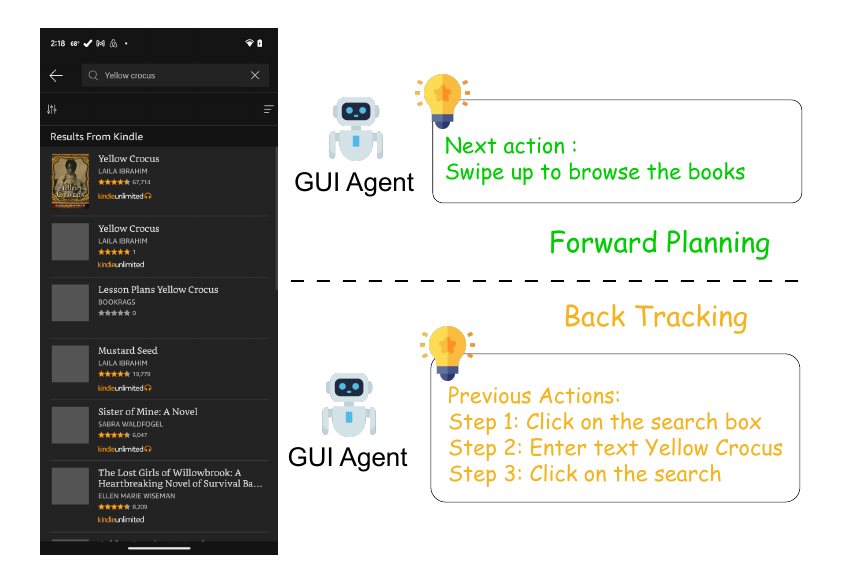}
\vspace{-0.5em}
\caption{Difference of forward-planning and back-tracking.}
\vspace{-0.5em}
\label{fig:intro}
\end{figure}  
\section{Related Works}
\subsection{Multimodal LLMs}
% 近期，闭源与开源的多模态大型语言模型（MLLM）不断发展，在非自然图像理解及 GUI 任务规划等领域均取得显著进展。从闭源模型来看，GPT-4V 和 GPT-4o 凭借强大视觉理解和任务规划能力，常作为 “大脑” 应用于计划和定位解耦的 Agent GUI Framework 。GPT-4V 融合文本与图像分析，打破传统大模型感知边界，是多模态 AI 重大进步；GPT-4o 相比前代，推理输出更高效，响应更快且成本更低。Gemini 模型系列涵盖多规模，其 Nano 模型虽小，在特定 GUI 任务中表现出色，为资源受限设备上的 GUI Agent 提供新选择。Claude 3.5 Sonnet 则开创仅视觉范式实现桌面任务自动化，不仅负责任务规划，还预测环境交互动作，充当完整的GUI Agent Model。
% 开源模型为GUI Agent的领域能力细化提供了可能。Qwen-VL系列以其细粒度的视觉理解和多模态能力而著称, 尤其是Qwen2-VL是过去绝对多数领域模型的MLLM基座模型。此外，现有开源模型在GUI代理领域展现出了各自独特的技术优势. InternVL-2的渐进式对齐策略和CogVLM的高效跨模态集成提升了多模态任务性能. Ferret通过空间理解优化了人机交互的精确性. LLaVA系列模型凭借模块化设计和数据效率加速了开发流程. BLIP-2的轻量化架构则为资源受限环境提供了高效解决方案。
Recent advances in closed-source and open-source multimodal large-language models (MLLMs) have significantly enhanced non-natural image understanding and GUI task planning. Closed-source models like GPT-4V~\cite{2023GPT4VisionSC} and GPT-4o~\cite{hurst2024gpt}, with strong visual and task-planning abilities, often serve as the ``brain'' in planning and grounding decoupled GUI agent framework. Open-source models have enabled domain-specific capability refinement for GUI Agents. The Qwen-VL~\cite{bai2023qwen,wang2024qwen2,bai2025qwen2} series distinguishes itself through fine-grained visual comprehension and multimodal capabilities, with Qwen2-VL~\cite{wang2024qwen2} notably serving as the foundational MLLM backbone for the majority of previous GUI Agent implementations. Other open-source models have distinct technical advantages in the GUI agent field. InternVL-2~\cite{chen2024far} improves multimodal task performance through progressive alignment. CogVLM~\cite{wang2024cogvlm} uses a visual expert module to fuse vision and language features. Ferret~\cite{you2023ferret} boosts human-computer interaction precision with enhanced spatial comprehension. The LLAVA~\cite{liu2023visual,liu2024improved,li2024llava} series is used for its lightweight projection layer, enabling fast training and multimodal understanding.

\subsection{GUI Agents}
% 目前流行的GUI Agen主要基于MLLMs， 旨在跟随人类指令完成在手机或者电脑的自主代理操作，涵盖Web 、Desktop 和Mobile 等应用场景。其完成代理的过程可以抽象为planning 和 grounding两个阶段。按照两个阶段是否由独立的模型完成，现有的GUI Agent作品可以分为单纯的GUI Grounding Model(如[Aria-UI][Uground]) 和统一的GUI Agent framework (如[aguvis][os-atlas][ui-tars])。
% 从感知内容的来看，早期作品 [WebPilot] [Hybrid Agent] [WebDreamer] [Agent Q][LASER][SeeAct] 聚焦于 in Web platforms for automating web interaction and navigation. They integrated structured text (e.g., accessibility trees, HTML-DOM) with environmental visual structures (screenshots), establishing a solid foundation in GUI Agent fields. However, the difficulty of accessing structured text in real-world settings like desktop and iOS applications, coupled with its token inefficiency impacting MLLM performance, drove a shift toward vision-only paradigms. Claude 3.5 Sonnet[] (Computer Use) pioneered this paradigm in desktop task automation, unifying task planning and environmental interaction action prediction into one system. 
% The rise of vision-only methods also sparked cross-platform unified modeling. Recent methods such as AGUIVS[], UI-TARS[], os-atlas[] 使用 uniform action space, 并在collected multi-platform datasets 上进行训练。
% aguvis 将明确的计划和推理整合在其框架之中，增强了与复杂数字环境进行交互的能力。OS-ATLAS提出了一个标准化跨平台数据集的统一动作空间，包括基本动作和定制动作。UI-TARs将多种推理范式使用到了模型的planning阶段，包括任务分解，反思，里程碑式识别。
Recent GUI agents, predominantly built on MLLMs, can execute autonomous operations on phones or computers as per human instructions, applicable in web, desktop, and mobile scenarios. Their operation involves planning and grounding phases. Based on whether these two phases are handled by separate models, existing GUI agent works can be categorized into pure GUI grounding models like Aria-UI~\cite{yang2024aria} and Uground~\cite{gou2024navigating}, and unified GUI agent frameworks such as Aguvis~\cite{xu2024aguvis}, OS-ATLAS~\cite{wu2024atlas} and UI-TARS~\cite{qin2025ui}.

In terms of perceptual content, early works such as WebPilot~\cite{zhang2025webpilot}, Hybrid Agent~\cite{song2024beyond}, WebDreamer~\cite{gu2024your}, Agent-Q~\cite{putta2024agent}, LASER~\cite{ma2023laser}, and SeeAct~\cite{zheng2024gpt} concentrated on web platforms to automate web interaction and navigation. They incorporated structured text (\textit{e.g.}, accessibility trees, HTML-DOM) with environmental visual structures (screenshots), establishing a foundation in the GUI agent field. However, the difficulty of accessing structured text in real-world settings like desktop and iOS applications, along with its token inefficiency impacting MLLM performance, has driven a shift toward vision-only approaches. Claude 3.5 Sonnet (Computer Use)~\cite{hu2024dawn}pioneered this paradigm in desktop task automation, integrating task planning and environmental interaction action prediction into a single system.

The rise of vision-only methods has also sparked cross-platform unified modeling. Recent methods like Aguvis~\cite{xu2024aguvis}, UI-TARS~\cite{qin2025ui}, and OS-ATLAS~\cite{wu2024atlas} use a uniform action space and are trained on multi-platform datasets. Aguvis integrates explicit planning and reasoning into its framework, enhancing interaction with complex digital environments. OS-ATLAS proposes a unified action space for standardized cross-platform datasets, covering basic and custom actions. UI-TARS apply diverse reasoning patterns in the model's planning phase, including task decomposition, reflection, and milestone recognition.

% The lightweight framework of BLIP-2[] delivers efficient solutions for resource-constrained environments.

%  GPT-4V integrates text and image analysis, transcending traditional LLM perception boundaries and marking a major multimodal AI breakthrough. GPT-4o improves on its predecessor with more efficient reasoning, faster responses and lower costs.

% The Gemini model family[], available in various scales, includes the Nano model, which, despite its size, excels in specific GUI tasks, offering a new option for GUI Agents on resource constrained devices. 
\section{Method}
\label{sec:Method}
%ScaleTrack的优化分为两个阶段：接地和规划。在\ref{sec: Data_Scale}节中，我们介绍了接地阶段的data scale。在\ref{sec: back-tracking}节中，我们将讨论规划阶段的回溯和规划。
The optimization of ScaleTrack is divided into two stages: grounding and planning. In Section~\ref{sec: Data_Scale}, we introduce the data scale in the grounding stage. In Section~\ref{sec: back-tracking}, we discuss forward-planning and back-tracking in the Planning stage. 

\subsection{Formulation}
\label{sec: PROBLEM}
Given a task description and the initial environment observation $o_1$, the autonomous GUI carries out planning and makes a prediction of an action that belongs to the action space, namely $a_1 \in \mathbb{A}$. Subsequently, the client side environment is updated upon receiving this action and provides a new observation $o_2$. The above process is repeated continuously until the predicted action is \textit{terminate}, which signifies the completion of the task. The entire process can be format as:
\begin{equation}\label{eq_Future}
    a_n = \mathcal{M}_{\theta}(\text{task}, (o_1,a_1), ... , (o_{n-1},a_{n-1}),o_n),
\end{equation}
where $\mathcal{M}$ is the policy, equivalent to the GUI agents model, and $\theta$ denotes its associated parameters. 

\begin{figure*}[t]
\centering
\includegraphics[width=1.0\textwidth]{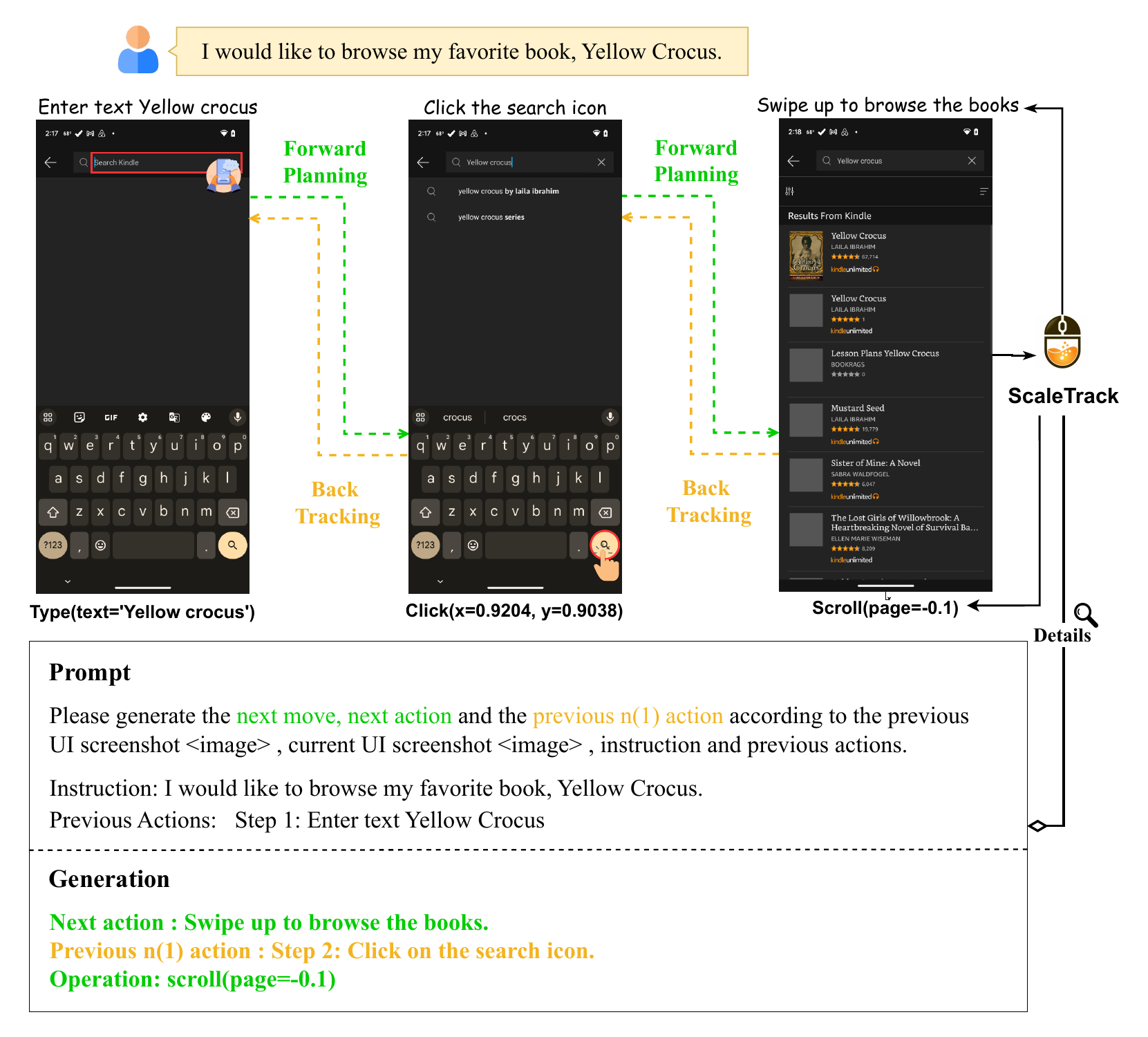}
\vspace{-1em}
\caption{Overall description of our proposed ScaleTrack in processing task instruction and generating actions via forward-planning and back-tracking, as well as the format of training data}
\vspace{-1em}
\label{fig:overview}
\end{figure*}

\subsection{Data Scale Across Different Domains and Types}
\label{sec: Data_Scale}

\subsubsection{Unified Data Format} 
\label{sec: Unified}
%一个重要的问题是vlm如何表示图像中的数值坐标。我们将实际坐标映射到0-1000的相对坐标当中，再进行缩放，最终得到一个0-1之间的数值用于表示相对于图像大小的距离。 使用相对坐标的好处是对于不同分辨率的图像具有统一的表示，在保持相同长宽比的情况下也可进行自由的resize操作以适应不同输入token长度限制的模型。
% hwk -> 如何表示GUI screenshot中的文字、图标、控件的空间信息是GUI Agent model首要需要解决的问题。提出的ScaleTrack将实际坐标映射到0-1000的相对坐标当中，再进行缩放，最终得到一个0-1之间的数值用于表示相对于图像大小的距离。使用相对坐标的好处是对于不同分辨率的图像具有统一的表示，在保持相同长宽比的情况下也可进行自由的resize操作以适应不同输入token长度限制的模型。

%其次，在接地任务中，坐标表示通常采用方框形式：$(x1, y1, x2, y2)$，其中$x1, y1, x2,$和$y2$分别表示包含对象的最小边界框的左上角和右下角的横坐标和纵坐标。然而，在代理任务中，还可以采用点$(x, y)$的形式表示需要点击的位置，其中$x$和$y$分别表示要单击的点的水平和垂直坐标。因此，我们统一将来自各个源的数据统一成point的形式。
% hwk -> 过去的GUI定位数据集的标注通常采用方框形式：$(x1, y1, x2, y2)$，其中$x1, y1$和$x2, y2$分别表示包含对象的最小边界框的左上角和右下角的坐标。然而在诸多 GUI Agent 操作中，基于click的交互更为普遍。例如，用户点击 GUI 界面的按钮或链接时，本质上就是点操作。因此，我们将坐标表示统一为点格式，可更好地契合 GUI Agent 任务的操作需求，提升元素定位与交互的精准度和效率。
% An important question is how VLMs represents numerical coordinates in images. In this work, we map the absolute coordinates to relative coordinates of 0 - 1000, then scale them, and finally get a value between 0 - 1 to represent the distance relative to the image size. The advantage of using relative coordinates is that images of different resolutions have a unified representation. Moreover, while maintaining the same aspect ratio, they can also be freely resized to adapt to models with different input token length limits.

How to represent the spatial information of texts, icons and controls in GUI screenshots is the primary issue that the GUI Agents need to address. The proposed ScaleTrack maps actual coordinates to relative coordinates ranging from 0 to 1000, then scales them to obtain a value between 0 and 1 to indicate the distance relative to the image size. The advantage of using relative coordinates is that they provide a unified representation for images of different resolutions and allow for free resizing while maintaining the same aspect ratio, thus catering to models with varying input token length constraints.

% Secondly, in grounding tasks, the coordinate representation is typically in the form of a box: $(x1, y1, x2, y2)$, where $x1, y1, x2,$ and $y2$ denote the horizontal and vertical coordinates of the upper left and lower right corners of the minimal bounding box that contains the object, respectively. However, in agent tasks, the location to be clicked can also be represented in the form of a point $(x, y)$, where $x$ and $y$ represent the horizontal and vertical coordinates of the point to be clicked, respectively.

In the previous GUI grounding dataset annotation, the coordinates of the target element are typically in the form of a box: $(x1, y1, x2, y2)$, where $(x1,y1)$ and $(x2,y2)$ indicate the coordinates of the top-left and bottom-right corners of the smallest bounding box enclosing the element. However, in many GUI Agent operations, click-based interaction is more prevalent. For instance, when users click buttons or links on the GUI, it is essentially a point operation. Hence, we unify the coordinate representation into the point format so that it can better suit the operational requirements of GUI Agents tasks, thereby enhancing the accuracy and efficiency of element grounding and interaction.

\subsubsection{Merge Data Sources} 
\label{sec: Merge}
%早期的GUI Agent通过将屏幕描述成结构化的文本或直接获取屏幕的源数据，利用纯文本的大语言模型进行任务规划和环境感知。自Seeclick \cite{cheng2024seeclick}以来，许多工作尝试使用屏幕截图的纯视觉作为输入，从而实现不同接口、平台之间的泛化。然而，和计算机视觉领域内积累的大量通用图像数据相比，GUI Agent领域涉及的Grounding数据还存在很大的局限性：不同平台、不同设备的数据难以泛化；不同的工作使用孤立的数据合成准则，难以形成互补。 

%为了解决GUI接地中的泛化问题，我们对GUI agent领域不用来源、不同合成准则的数据进行了融合。 具体来说，following Uground \cite{}，引入了通过综合多种参考模式，构建的大量的website接地数据； following Aria-UI \cite{}，引入了通过综合上下文感知构造的Context-aware的接地数据；inspired by Augvis \cite{}，引入了通过统一动作空间构造的指令型的Grounding数据。我们的工作将这些数据统一了起来，研究不同数据构造方式之间是否可以互相补充，从而集成从各个方面提高Agent 定位泛化能力的数据。
In early work, agents interact with the environment by extracted structured data (\textit{e.g.}, HTML-DOM), and use large language models in plain text for task planning. Since SeeClick~\cite{cheng2024seeclick}, many works have tried to use the pure vision of screenshots as input to achieve generalization between different interfaces and platforms. However, compared with the large amount of general image data accumulated in the field of computer vision, the grounding data involved in the GUI Agents field still has great limitations: data from different platforms and devices are difficult to generalize and different tasks use isolated data synthesis criterion, which makes it difficult to complement each other.

In order to solve the generalization problem in GUI grounding, we fused data from different sources and with different synthesis criterion in the field of GUI agents. Specifically, following Uground~\cite{gou2024navigating}, we introduced a large amount of website grounding data constructed by integrating multiple reference modes. Following Aria-UI~\cite{yang2024aria}, we introduced context-aware grounding data constructed by integrating context perception. Inspired by Aguvis~\cite{xu2024aguvis}, we introduced template-enhanced grounding data constructed by a unified action space. As shown in Table~\ref{tab-data}, our work unifies these data and studies whether different data synthesis criterion can complement each other, thereby improving the generalization ability of Agent positioning from all aspects.

% data
\begin{table*}[ht]
\centering
\caption{Statistics of the open-source grounding and planning data.}
\label{tab-data}
\begin{adjustbox}{max width=2.0\columnwidth}
\begin{tabular}{@{}lcccccc@{}}
\toprule
\multirow{2}{*}{\textbf{Data Source}} & \multirow{2}{*}{\textbf{Data Type}} & \multicolumn{2}{c}{\textbf{Grounding}} & \multicolumn{2}{c}{\textbf{MultiStep}} \\ \cmidrule{3-6}
 & & \textbf{Elements} & \textbf{Screenshots} & \textbf{Trace} & \textbf{avg steps} \\ \midrule
\multirow{1}{*}{Uground} & Web & 9M & 773K  & /  &  / \\ \midrule
 \multirow{3}{*}{OS-Atlas} & Web &7.8M & 1.6M & / & / \\
 & Mobile &1.1M &107K & / & / \\
 & Desktop & 11.3M &54K & / & / \\ \midrule
 \multirow{3}{*}{Aria-UI} & Web &- & 173K &/ &/ \\
 & Mobile &- &104K &/ &/ \\
 & Desktop &- &7.8K &/ &/ \\ \midrule
 \multirow{3}{*}{Aguvis} & Web & 723K & - & 6.3k & 6.7 \\
 & Mobile & 232K & - & 28.7K & 8.5 \\
 & Desktop & 7K & - & / & / \\ \midrule
Ours & Ours & * & 7.5M & * & 8.2 \\ \bottomrule
\end{tabular}
\end{adjustbox}
\end{table*}

%As summarized in Table \ref{}, the main works currently engaged in exploration and open-sourcing on data aspects include OS-Atlas\cite{wu2024atlas}, Uground\cite{gou2024navigating}, Aguvis\cite{xu2024aguvis}, and Aria-UI\cite{yang2024aria}. To investigate the impact of data scale on grounding capabilities, we collected and unified data from various sources, resulting in approximately 4,330,734 grounding data points. We then conducted data scaling experiments on this dataset.

%在现实世界的场景中，一个复杂的屏幕截图可能包含数百个UI元素，因此常规的数据集自然会在一个图像中包含多个对象。为了减少训练过程中反复加载图像的冗余操作，节省训练成本，我们将同一截图的不同接地指令合并到一个对话集中，从而创建多回合对话训练数据。
In real-world scenarios, a complex screenshot may contain hundreds of UI elements, and existing open-source grounding datasets naturally include multiple objects within a single image. To avoid redundant operations of repeatedly loading images and to reduce training overhead, we merged different instruction/description-answer pairs from the same screenshot into a single conversation, thus creating multi-turn training data.

\subsection{Enhancing Action Understanding with back-tracking}
\label{sec: back-tracking}

\subsubsection{From Predicting the Future to back-tracking} 
\label{sec: subback-tracking}
%如\ref{sec: PROBLEM}所述，以往的GUI agent planning阶段的训练往往将agent与环境的交互建模为部分可观察的马尔可夫决策过程，为模型提供过去的行为和状态感知， 要求模型对未来的动作概率进行预测。这个过程忽略了模型对历史决策的反思和回溯，即模型只知道自己到达了什么样的状态，却不知道自己是如何到达该状态的。为此，我们将agent与环境的交互进行扩展，在每个时间步t，模型不仅仅要预测当前整体目标下的下一步动作，还要预测自己是如何达到当前状态的（即历史时刻的动作）。具体来说，在时间步t，模型的预测任务定义如下：
%与这些方法不同，我们提出了一个新的任务，要求模型不仅要计划未来，还要预测过去。形式上，现有工作中模型的任务定义如下：
% Unlike these approaches, we propose a new task that requires the model to not only plan for the future but also predict the past. Formally, the task definition for models in existing work is as follows:
% \begin{equation}\label{eq_back-tracking}
%     a_{n-1},a_n = \mathcal{M}_{\theta}(\text{task}, (o_1,a_1), ... , (o_{n-1}), o_n)
% \end{equation}
As mentioned in Section~\ref{sec: PROBLEM}, the training of GUI agents planning stage is usually modeled as a partially observable Markov decision process. It provides the model with past behaviors and state perceptions, only requiring the model to perform forward-planning and predict the probabilities of future actions based on these. However, this approach overlooks the model's ability to reflect on and backtrack historical decisions. That is, the model is aware of the state it has reached but is unaware of how it arrived there.
To address this limitation, we extend the interaction between GUI agents and their environment by incorporating back-tracking. Specifically, at each time step t, ScaleTrack not only predicts the next action under the current overall goal but also predicts the historical actions that led to the current state. This can be formulated as follows:
\begin{equation}
\label{eq_back-tracking}
    a_{n-1},a_n = \mathcal{M}_{\theta}(\text{task}, (o_1,a_1), ... , (o_{n-1},a_{n-1}), o_n)
\end{equation}
In the forward-planning aspect, similar to traditional methods, ScaleTrack takes the current state and task instructions as input and generates the next action probability distribution. This enables the agents to determine the most likely next action to perform in the current state, thereby achieving step-by-step task execution.

In terms of back-tracking, ScaleTrack introduces a reverse prediction mechanism. Based on the current state and task instructions, it predicts the actions that might have occurred prior to the current state. By doing so, the agents can gain a clearer understanding of the path it took to reach the current state, enabling it to better assess the rationality of its previous actions and adjust its subsequent planning accordingly.

\subsubsection{Different Expressions of Previous Actions} 
\label{sec: Previous_actoion}
State observation achieves the conversion from structured data such as HTML to screenshots. Previous actions also have two different forms of expression: a low-level instruction in natural language and an actual sequence of operational actions (\textit{e.g.}, the text annotations in the top and bottom of each frame in Fig.~\ref{fig:overview}). To study the impact of different expressions of previous actions on the VLM's understanding of historical actions, we conducted experiments in both the training and testing phases.

% Specifically, examples of historical action sequences in the two different forms of expression are as shown in~\ref{fig:overview}.

% \subsection{Training Pipline}

\section{Experiments}
\label{sec:experiments}

In this section, we conduct experiments on GUI Grounding Evaluation and Offline/Online GUI Agent Evaluation, respectively. We chose Qwen2-VL-7B\cite{wang2024qwen2} as the base model for training and fine-tuned it using the data introduced in Section~\ref{sec:training data}. The training is divided into two steps: Grounding and Planning with back-tracking.

%%%%%%%%%%%%%%%%%%%%%%%%%%%%%%%%%%%%%%%%%%%
\subsection{Training Details}
\label{sec:training details}
% 为了评估数据scale对agent grounding能力的影响，我们在ScreenSpot数据集上进行了评估。我们首先对比了我们的模型和其他开源模型的效果，然后评估了不同数据规模下模型grounding能力的变化。
%表1的结果表明，ScaleTrack在不同的平台上均表现出优秀的GUIgrounding能力。实验结果表明，通过统一不同来源的开源数据进行统一微调，
\subsubsection{Training Data} 
\label{sec:training data}
%对于Grounding阶段的训练，我们集成了开源数据：OS-Atlas\cite{wu2024atlas}, Uground\cite{gou2024navigating}, Aguvis\cite{xu2024aguvis}, and Aria-UI\cite{yang2024aria}，并将它们标准化为我们统一的动作空间格式。我们在表2中提供了训练的基础数据统计。
For training in the grounding stage, we integrated open-source data: OS-Atlas~\cite{wu2024atlas}, Uground~\cite{gou2024navigating}, Aguvis~\cite{xu2024aguvis}, and Aria-UI~\cite{yang2024aria}, and standardized them into the unified format. We provide basic data statistics for training in Table~\ref{tab-data}. For training in the Planning stage, we selected Aguvis~\cite{xu2024aguvis}, a dataset annotated with Observation, Thought, and Low-level Instruction as the data source, and performed back-tracking transformations. In addition, in order to study the impact of the format of previous actions on the model's understanding of action sequences, the previous actions of the data were replaced to obtain a copy.
\subsubsection{Training Settings} 
\label{sec: Training Settings}
We choose Qwen2-VL\citep{wang2024qwen2} as the base model for training and employ the AdamW optimizer with a learning rate of $1e{-5}$ and employ a cosine learning rate scheduler with a warm-up ratio of $0.03$ steps. We utilize a global batch size of 128 in the grounding stage and 64 in the planning stage and employ DeepSpeed ZERO3-style data parallelism. We train ScaleTrack following a two-stage procedure. First, all grounding data is used to train ScaleTrack's basic GUI grounding capability. Then, based on the model trained in the grounding stage, planning data with forward-planning and historical back-tracking are input into the model to further enhance the planning ability of the model. We train ScaleTrack on a cluster of 4 nodes V100-80G GPUs.

%%%%%%%%%%%%%%%%%%%%%%%%%%%%%%%%%%%%%%%%%%%
\subsection{Grounding Capability Evaluation}
\label{sec:experiments-grounding evaluation}
% 为了评估数据scale对agent grounding能力的影响，我们在ScreenSpot数据集上进行了评估。我们首先对比了我们的模型和其他开源模型的效果，然后评估了不同数据规模下模型grounding能力的变化。
In order to evaluate the impact of data scaling on the agent's grounding ability, we conducted experiments on the ScreenSpot dataset. We first compared the accuracy of our model with other baselines, and then evaluated the changes in the model grounding ability under different data scales.
%表\ref{tab-1}报告了我们提出的ScaleTrack和各种基线的准确性分数，包括开源模型，如UGround \cite{}, Aria-UI \cite{}, OS-Atlas \cite{}, Aguvis \cite{}和uitars \cite{}，它们使用私人数据。从表中可以看出，ScaleTrack的性能明显超过了基线方法，达到了最先进的水平。结果证明了数据扩展所获得的性能。相比于任何一种孤立的数据生成策略，ScaleTrack使用的综合数据生成模式都取得了一定的改进，该方法的优越性还体现了将不同来源的数据结合起来的优势。

\textbf{ScreenSpot.} Table~\ref{tab-1} reports the accuracy scores of our proposed ScaleTrack and various baselines, including the open source models such as UGround\cite{gou2024navigating}, Aria-UI~\cite{yang2024aria}, OS-Atlas\cite{wu2024atlas}, Aguvis~\cite{xu2024aguvis} and UI-TARS~\cite{qin2025ui}, which uses In-house data. We can see from the table that the performance of ScaleTrack clearly surpasses those of the baseline methods that use open-source data and outperforms the previous state-of-the-art(SOTA) model \cite{xu2024aguvis} by 1.2\% in the average Score. Moreover, ScaleTrack gains a more obvious advantage in Icon/Widget sub-items that are more difficult to generalize, with improvements of 4.4\%, 8.6\%, and 7.8\% respectively. The results demonstrate the generalization ability gained by the data scaling. The comprehensive data synthesis strategy used by ScaleTrack achieves better results than any isolated data synthesis criterion, and the superiority of our method also shows the advantage of combining data from different sources.
%为了进一步分析接地数据缩放的有效性，我们绘制了ScaleTrack在ScreenSpot上不同训练步骤的准确率分数。如图\ref{}所示，随着数据规模的扩大，平均准确率得分会提高，这证明了GUI agent数据 Scaling 的现象依旧存在。

% grounding
\begin{table}[ht]
\centering
\caption{Comparison of various planners and grounding methods on ScreenSpot.}
\label{tab-1}
\begin{adjustbox}{max width=1.0\columnwidth}
\begin{tabular}{@{}lllccccccc@{}}
\toprule
\multicolumn{2}{l}{\multirow{2}{*}{\textbf{Method}}} & \multirow{2}{*}{\textbf{Data Source}} & \multicolumn{2}{c}{\textbf{Mobile}} & \multicolumn{2}{c}{\textbf{Desktop}} & \multicolumn{2}{c}{\textbf{Web}} & \multirow{2}{*}{\textbf{Avg}} \\ \cmidrule{4-9} 
 & & & \textbf{Text} & \textbf{Icon/Widget} & \textbf{Text} & \textbf{Icon/Widget} & \textbf{Text} & \textbf{Icon/Widget} & \\ \midrule 
\multicolumn{10}{l}{\textit{Agent Framework}} \\ \midrule

\multirow{3}{*}{GPT-4} & SeeClick &Public &76.6 &55.5 &68.0 &28.6 &40.9 &23.3 &48.8 \\
 & OmniParser &In-house &93.9 &57.0 &91.3 &63.6 &81.3 &51.0 &73.0 \\
 & UGround-7B &Public &90.1 &70.3 &87.1 &55.7 &85.7 &64.6 &75.6 \\
\noalign{\vskip 2pt} % 上方增加空白
\hdashline
\noalign{\vskip 2pt} % 下方增加空白
\multirow{2}{*}{GPT-4o} & SeeClick & Public &81.0 &59.6 &69.6 &33.6 &43.9 &26.2 &52.3 \\
 & UGround-7B & Public &93.4 &76.9 &92.8 &67.9 &88.7 &68.9 &81.4 \\
\midrule
\multicolumn{10}{l}{\textit{Agent Model}} \\ \midrule
\multicolumn{2}{l}{GPT-4o} & In-house &20.2 &24.9 &21.1 &23.6 &12.2 &7.8 &18.3 \\
\multicolumn{2}{l}{Claude Computer Use} & In-house &- &- &- &- &- &- &83.0 \\
\multicolumn{2}{l}{Gemini 2.0 } & In-house &- &- &- &- &- &- &84.0 \\
\multicolumn{2}{l}{UI-TARS-7B} & In-house &94.5 &85.2 &95.9 &85.7 &90.0 &83.5 &89.5 \\ 
\noalign{\vskip 2pt} % 上方增加空白
\hdashline
\noalign{\vskip 2pt} % 下方增加空白
\multicolumn{2}{l}{CogAgent} & Public &67.0 &24.0 &74.2 &20.0 &70.4 &28.6 &47.4 \\
\multicolumn{2}{l}{SeeClick} & Public &78.0 &52.0 &72.2 &30.0 &55.7 &32.5 &53.4 \\
\multicolumn{2}{l}{Qwen2-VL} & Public &75.5 &60.7 &76.3 &54.3 &35.2 &25.7 &55.3 \\
\multicolumn{2}{l}{UGround-7B} & Public &82.8 &60.3 &82.5 &63.6 &80.4 &70.4 &73.3 \\
% \multicolumn{2}{l}{AGUVIS-G-7B} & Public &95.6 &77.7 &93.8 &67.1 &88.3 &75.2 &84.4 \\
\multicolumn{2}{l}{OS-Atlas-7B} & Public &93.0 &72.9 &91.8 &62.9 &90.9 &74.3 &82.5 \\
\multicolumn{2}{l}{AGUVIS-7B} & Public &95.6 &77.7 &93.8 &67.1 &88.3 &75.2 &84.4 \\
\midrule
\multicolumn{2}{l}{ScaleTrack-7B} & Public &93.8  &82.1  &91.7 &75.7 &87.4 &83.0  &86.8\\  

\bottomrule 
\end{tabular}
\end{adjustbox}
\end{table}

\begin{figure}[t]
\centering
\includegraphics[width=0.9\textwidth]{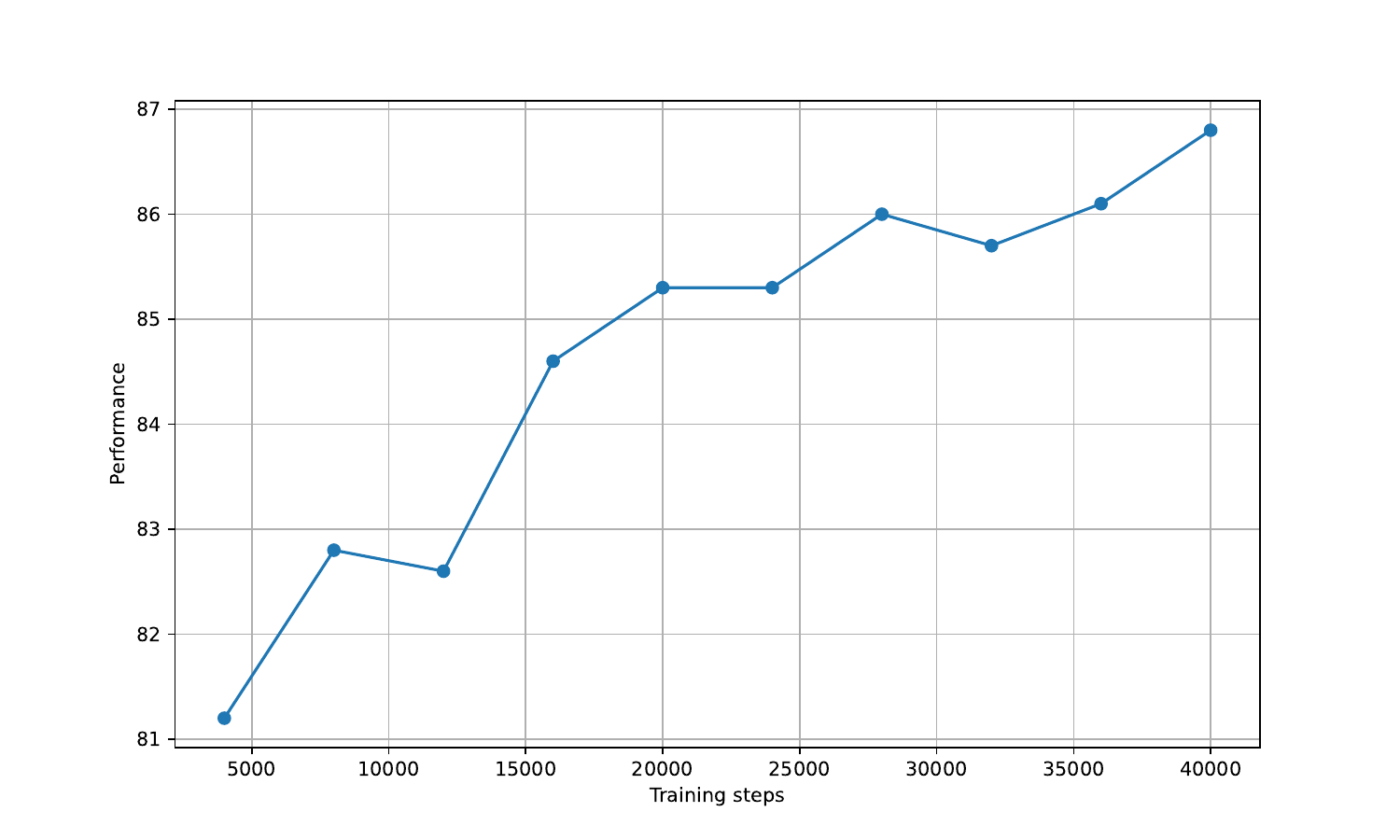}
\vspace{-0.5em}
\caption{Scaling curve of ScaleTrack-7B on ScreenSpot.}
\vspace{-0.5em}
\label{fig:scale}
\end{figure}

\textbf{The Effect of Grounding Data Scaling.} To further analyze the effectiveness of grounding data scaling, we plot the accuracy scores of ScaleTrack on ScreenSpot in the different training steps. As illustrated in Figure~\ref{fig:scale}, as the data scales up, the average accuracy score fluctuates, but overall it will gradually improve. The result suggests great potential for continuously expanding grounding data to improve performance.

%图2的结果表明，随着数据的scaling，模型的GUI grounding能力在逐步提高

\subsection{Offline Agent Capability Evaluation}
\label{sec:experiments-offline evaluation}
%我们使用Android自动化数据集AndroidControl在移动设备上评估ScaleTrack，该数据集包含15,000个演示，来自人类评分员在Android设备上的833个不同应用程序（跨越40个应用类别）上执行各种各样的任务。我们随机抽样500个任务来创建一个子集，其中包括\cite{}步骤。我们报告了在高级和低级任务中对域外数据的操作类型准确性、接地准确性和步骤成功率。
\textbf{AndroidControl.} We evaluate ScaleTrack on mobile devices using Android automation dataset  AndroidControl, which encompasses 15,000 demonstrations from human raters performing a diverse variety of tasks on 833 different apps spanning 40 app categories on Android devices. Following~\cite{li2024effects,xu2024aguvis}, we randomly sample 800 steps to create a subset. We report the action type accuracy, grounding accuracy, and step success rate on out-of-domain data within both high-level and low-level tasks. The evaluation process of AndroidControl-High relies on historical action input, and we strictly follow OS-Atlas \cite{wu2024atlas} to use low-level natural language to describe the historical actions.

%如表\ref{tab-AndroidControl}所示，ScaleTrack超越了使用低水平和高水平可用数据的强基线，低水平和高水平的步进成功率分别为{86.6％}和{77.9％}。值得注意的是ScaleTrack planning阶段的数据来源和Aguvis相同，并没有添加额外的Planning数据标注，这证明了我们的back-tracking策略的有效性和可扩展性。 需要注意的是，对于AndroidControl-High数据集，我们发现使用不同的previous action
As depicted in Table \ref{tab-AndroidControl}, ScaleTrack surpassed strong baselines that use available data on both Low and High levels, achieving the step success rate of {86.6\%} for the Low level and {77.9\%} for the High level. Notably, the data source of the ScaleTrack's planning stage is the same as that of Aguvis, and no additional planning data annotations are added, which proves the effectiveness and scalability of our back-tracking strategy. 

\textbf{GUI-Odyssey.} GUI Odyssey is a comprehensive dataset for evaluating cross-app navigation agents. It consists of 7,735 episodes from 6 mobile devices.  Following~\cite{xu2024aguvis}, we randomly sample 500 episodes to create a subset and report the action type accuracy, grounding accuracy, and step success rate. Table~\ref{tab-Odyssey} reports that ScaleTrack-7B achieved the best performance among the public data models on step success rate.
% offline
\begin{table}[ht]
\centering
\caption{Comparative results on AndroidControl dataset with two settings (AndroidControl-Low and -High).}
\label{tab-AndroidControl}
\begin{adjustbox}{max width=0.9\textwidth} % 修正宽度参数
\begin{tabular}{@{}llcccccc@{}}
\toprule
\multirow{2}{*}{\textbf{Agent Models}} & \multirow{2}{*}{\textbf{Data Source}} & \multicolumn{3}{c}{\textbf{AndroidControl-Low}} & \multicolumn{3}{c}{\textbf{AndroidControl-High}} \\ \cmidrule(l){3-8} &
& \textbf{Type} & \textbf{Grounding} & \textbf{SR} & \textbf{Type} & \textbf{Grounding} & \textbf{SR} \\ \midrule
Claude &In-house &74.3 &0.0 &19.4 &63.7 &0.0 &12.5 \\
GPT-4o &In-house &74.3 &0.0 &19.4 &66.3 &0.0 &20.8 \\
InternVL-2-4B &In-house &90.9 &84.1 &80.1 &84.1 &72.7 &66.7 \\
Qwen-2VL-7B &In-house &91.9 &86.5 &82.6 &83.8 &77.7 &69.7 \\
UI-TARS-7B &In-house &98.0 &89.3 &90.8 &83.7 &80.5 &72.5 \\
\noalign{\vskip 2pt} % 上方增加空白
\hdashline
\noalign{\vskip 2pt} % 下方增加空白
SeeClick &Public &93.0 &73.4 &75.0 &82.9 &62.9 &59.1 \\
Aria-UI &Public &- &87.7 &67.3 &- &43.2 &10.2 \\
OS-Atlas-4B &Public &91.9 &83.8 &80.6 &84.7 &73.8 &67.5 \\
OS-Atlas-7B &Public &93.6 &88.0 &85.2 &85.2 &78.5 &71.2 \\
Aguvis-7B &Public &- &- &80.5 &- &- &61.5 \\
\midrule
ScaleTrack-7B &Public & 93.9 & 84.9 & 86.6 &  89.2 &  72.8 &  77.9 \\

\bottomrule
\end{tabular}
\end{adjustbox}

% \begin{tablenotes}
%      \footnotesize
%      \item[*] *The evaluation process of AndroidControl-High relies on historical action input, and we strictly follow OS-Atlas \cite{wu2024atlas} to use low-level natural language to describe the historical actions.
% \end{tablenotes}
\end{table}

\begin{table}[ht]
\centering
\caption{Comparative results on GUI Odyssey dataset.}
\label{tab-Odyssey}
\begin{adjustbox}{max width=0.9\textwidth}
\begin{tabular}{@{}llccc@{}}
\toprule
\multirow{2}{*}{\textbf{Agent Models}} & \multirow{2}{*}{\textbf{Data Source}} & \multicolumn{3}{c}{\textbf{GUI Odyssey}} \\ \cmidrule(l){3-5}
& & \textbf{Type} & \textbf{Grounding} & \textbf{SR} \\ \midrule
Claude$^*$ &In-house &60.9 &0.0 &3.1 \\
GPT-4o &In-house &34.3 &0.0 &3.3 \\
InternVL-2-4B &In-house &82.1 &55.5 &51.5 \\
Qwen-2VL-7B &In-house &83.5 &65.9 &60.2 \\
UI-TARS-7B &In-house &94.6 &90.1 &87.0 \\
\addlinespace % 添加额外行间距
\hdashline
\addlinespace % 添加额外行间距
SeeClick &Public &71.0 &52.4 &53.9 \\
Aria-UI &Public &- &86.8 &36.5 \\
OS-Atlas-4B &Public &83.5 &61.4 &56.4 \\
OS-Atlas-7B &Public &84.5 &67.8 &62.0 \\
Aguvis-7B &Public &- &- &- \\
\midrule
ScaleTrack-7B &Public &85.6 &69.3 &65.3 \\
\bottomrule
\end{tabular}
\end{adjustbox}
\end{table}

% 值得注意的是，我们发现在high level子集的测试中，以往的工作在组织输入的时候似乎存在分歧，有些工作如aguvis使用类似‘click', 'type'等实际的动作序列作为历史动作的描述。 而某些工作如OS-Atlas 则使用low-level的instruction如‘Click on the question how to cancel or change my reservation?’ 作为历史动作的描述格式。 这种差异对于模型理解历史动作序列存在不同的影响， 为了促进评测流程的统一和该领域的研究，我们分别使用两种方式都进行了测试，并对结果进行了详细的分析和比较。
\textbf{The impact of action description in AndroidControl.} It is worth noting that we discovered a divergence in organizing inputs during testing on the AndroidControl's high-level subset in previous work. Some works, such as Aguvis~\cite{xu2024aguvis}, seem to use the action sequences(i.e. 'click' and 'type') as the description of historical actions. In contrast, other works, such as OS-Atlas~\cite{wu2024atlas}, use low-level instructions(i.e. 'Click on the question how to cancel or change my reservation?') as the description of historical actions. This difference has an impact on the model's understanding of historical action sequences. In order to promote the consistency of the evaluation process and research in this field, we tested it using both descriptions and provided a detailed analysis and comparison of the results.

%我们在没有使用backing的数据上进行了实验。当训练和测试分别使用instruction 形式和action形式 的previous action时，训练和测试保持一致的setting将会得到较优的结果，否则将会发生性能损失。这说明对于GUI agent来说，一致的上下文动作表述方式是重要的，这对于planning阶段的data scale和数据标注具有重要的指导意义

We conducted experiments on the dataset without back-tracking. As shown in Table~\ref{tab:AndroidControl-pro}, when training and testing use instructions and action forms to describe the previous actions respectively, keeping the same settings for training and testing will get better performance, otherwise accuracy will decrease. This shows that for GUI agents, a consistent way of expressing contextual actions is important, which has crucial guiding significance for data scale and data annotation in the planning stage.

\begin{table}[htb]
\centering
\caption{The impact of different action description in AndroidControl-High dataset.}
\label{tab:AndroidControl-pro}
\begin{adjustbox}{max width=0.9\textwidth}
\begin{tabular}{@{}lccc@{}}
\toprule
\multirow{2}{*}{\textbf{Train-Test}} & \multicolumn{3}{c}{\textbf{AndroidControl-High}} \\ \cmidrule(l){2-4} 
& \textbf{Type} & \textbf{Grounding} & \textbf{SR}  \\ \midrule
Instruction-Action & 81.8 & 71.6 & 66.7   \\
% Instruction-Action & 81.0 & 70.3 & 64.0   \\
Action-Action & 89.2 & 75.2 & 77.4   \\
Action-Instruction & 82.3 & 66.2 & 68.3 \\
Instruction-Instruction & 88.3 & 73.7 & 76.1   \\
% Instruction-Instruction & 87.7 & 73.8 & 74.8   \\
\bottomrule
\end{tabular}
\end{adjustbox}
\end{table} 

\subsection{Online Agent Capability Evaluation}
\label{sec:ONLINE}
%为了更好地测试真实场景下模型的性能，我们还在实时交互基准上测试了ScaleTrack. 

To better test the performance of ScaleTrack in real-world environments, we also tested ScaleTrack on the real-time interaction benchmarks. We use AndroidWorld~\cite{rawles2024androidworld} and MobileMiniWob\cite{rawles2023androidinthewild} for online mobile agent evaluation in an Android emulator environment. We use GPT-4o as the planner and CaleTrack-7B to locate elements and instructions.

\textbf{AndroidWorld~\cite{rawles2024androidworld}} contains a highly reproducible benchmark of 116 hand-crafted tasks across 20 apps and calculates the final state success rate on the device by checking the final system state. As shown in Table~\ref{tab: AndroidWorld}, ScaleTrack achieves the highest average task success rate of 44\%, outperforming the baseline models, which further highlights that data-scaling and back-tracking help the model handle diverse element descriptions and instructions in real-world environments.

\textbf{MobileMiniWob~\cite{rawles2023androidinthewild}} containts 92 tasks from MiniWob++~\cite{zheng2023synapse}. As shown in Table~\ref{tab: AndroidWorld}, ScaleTrack outperforms existing work in task success rate on MobileMiniWobs when using GPT-4o as the planner, with an average SR of $44.0\%$ and $61\%$ on AndroidWorld and MobileMiniWob, respectively. This comparison particularly highlights the effectiveness of the data scaling and back-tracking in our GUI agents model. 

\begin{table}[ht]
\centering
\caption{Task Success Rates (SR) on AndroidWorld and MobileMiniWob.}
\label{tab: AndroidWorld}
\begin{adjustbox}{max width=1.0\textwidth} % 添加一个合适的选项
\begin{tabular}{@{}lccc@{}}
\toprule
\textbf{Planner} & \textbf{Grounding} & \textbf{AndroidWorld\_{SR}} & \textbf{MobileMiniWob\_{SR}}  \\ \midrule
GPT-4-Turbo & UGround & 31.0 & - \\
GPT-4o & UGround & 32.8 & - \\
GPT-4o & AGUVIS-7B & 37.1 & 55.0 \\ \midrule
GPT-4o & ScaleTrack-7B & 44 & 61.0  \\
\bottomrule
\end{tabular}
\end{adjustbox}
\end{table}

\subsection{Ablation Study}
\label{sec:Ablation Study}
%我们进一步研究ScaleTrack是如何通过回顾过去的历史动作从而提高模型的planning能力的，我们在三个offline 评测数据集上比较了加入back-tracking数据前后的模型性能。结果如表所示，在加入back-tracking数据后，在AndroidContorl-L、ndroidControl-H和GUI-Odyssey数据集上模型的准确率分别提高了，，，
We further studied how ScaleTrack improves the model's planning ability by tracking the history of actions that led to the current state. We compared the model performance before and after back-tracking data training on three offline evaluation datasets. 

% We show the ablation results in Table~\ref{tab-Ablation}. After training on back-tracking data, the accuracy of the model on AndroidControl-Low, AndroidControl-High, and GUI-Odyssey datasets increased by $1.8{\%}$, ${3.1\%}$ and ${0.2\%}$, respectively. The results demonstrate the performance gained by the back-tracking capability contribution.
We show the ablation results in Table~\ref{tab-Ablation}. After training on back-tracking data, the accuracy of the model on AndroidControl-High and GUI-Odyssey datasets increased by $1.8\%$ and $0.7\%$, respectively. Furthermore, the recognition accuracy of the model after back-tracking training in action type has increased by $1.8\%$, $0.9\%$ and $0.6\%$ respectively. The results demonstrate the action understanding and prediction abilities gained by the back-tracking training.

\begin{table}[]
\centering
\caption{Ablation results of ScaleTrack.}
\label{tab-Ablation}
\begin{adjustbox}{max width=0.9\width}
\begin{tabular}{@{}lccccccccc@{}}
\toprule
\multirow{2}{*}{\textbf{Agent Models}} & \multicolumn{3}{c}{\textbf{AndroidControl-Low}} & \multicolumn{3}{c}{\textbf{AndroidControl-High}} & \multicolumn{3}{c}{\textbf{GUI Odyssey}} \\ \cmidrule(l){2-10} 
& \textbf{Type} & \textbf{Grounding} & \textbf{SR} & \textbf{Type} & \textbf{Grounding} & \textbf{SR} & \textbf{Type} & \textbf{Grounding} & \textbf{SR} \\ \midrule
ScaleTrack-7B & 93.9 & 84.9 & 86.6 & 89.2 &  72.8 &  77.9 & 85.6 & 69.3 & 65.3 \\
w/o back-tracking & 92.1 & 85.6 & 86.6 & 88.3& 73.7 & 76.1 & 85 & 69.4 & 64.6 \\ \bottomrule
\end{tabular}
\end{adjustbox}
\end{table}

% \begin{table}[]
% \centering
% \caption{Ablation results of ScaleTrack.}
% \label{tab-Ablation}
% \begin{adjustbox}{max width=0.9\width}
% \begin{tabular}{@{}lccccccccc@{}}
% \toprule
% \multirow{2}{*}{\textbf{Agent Models}} & \multicolumn{3}{c}{\textbf{AndroidControl-Low}} & \multicolumn{3}{c}{\textbf{AndroidControl-High}} & \multicolumn{3}{c}{\textbf{GUI Odyssey}} \\ \cmidrule(l){2-10} 
% & \textbf{Type} & \textbf{Grounding} & \textbf{SR} & \textbf{Type} & \textbf{Grounding} & \textbf{SR} & \textbf{Type} & \textbf{Grounding} & \textbf{SR} \\ \midrule
% ScaleTrack-7B & 93.9 & 84.9 & 86.6 & 89.2 &  72.8 &  77.9 & 85.6 & 69.3 & 65.3 \\
% w/o back-tracking & 91.5 & 84.9 & 84.8 & 87.7 & 73.8 & 74.8 & 86.1 & 69.1 & 65.1 \\ \bottomrule
% \end{tabular}
% \end{adjustbox}
% \end{table}

\section{Conclusion}
In this work, we introduce ScaleTrack for scaling and back-tracking automated GUI agents.
We find two widespread problems including isolated data synthesis criterion and unconsidered back-tracking capability.
To alleviate these problems, we first propose to integrate several data-driven GUI element enhancement methods to scale the training process of GUI grounding, and unifies a wide range of grounding data into a fixed training template.
We then propose a hybrid training strategy to learn forward-planning and back-tracking capabilities simultaneously.
We conduct extensive experiments on several benchmark datasets like grounding evaluation, offline and online evaluation, and the results demonstrate the effectiveness of our proposed ScaleTrack.

\newpage
{
    \small
    \bibliographystyle{ieeenat_fullname}
    \bibliography{main}
}

\end{document}